\documentclass{article} 
\usepackage{iclr2027_conference,times}

\usepackage{amsmath,amsfonts,bm}
\usepackage[dvipsnames]{xcolor}

\newif\ifshowcomments
\showcommentstrue   

\def\eqref#1{equation~\ref{#1}}

\def\1{\bm{1}}

\def\rvx{{\mathbf{x}}}
\def\rvy{{\mathbf{y}}}
\def\rvz{{\mathbf{z}}}

\DeclareMathAlphabet{\mathsfit}{\encodingdefault}{\sfdefault}{m}{sl}
\SetMathAlphabet{\mathsfit}{bold}{\encodingdefault}{\sfdefault}{bx}{n}

\newcommand{\E}{\mathbb{E}}

\usepackage{hyperref}
\usepackage{url}

\usepackage{mathtools}
\usepackage{amssymb}

\usepackage{xcolor}
\usepackage{listings}
\usepackage{wrapfig}
\usepackage{algorithm}

\usepackage{booktabs,subcaption,adjustbox}
\usepackage[table]{xcolor}
\DeclareCaptionFont{ablation}{\AblationFont}

\usepackage{graphicx,colortbl,subcaption}
\def\eqref#1{equation~\ref{#1}}

\def\1{\bm{1}}

\def\rvx{{\mathbf{x}}}
\def\rvy{{\mathbf{y}}}
\def\rvz{{\mathbf{z}}}

\DeclareMathAlphabet{\mathsfit}{\encodingdefault}{\sfdefault}{m}{sl}
\SetMathAlphabet{\mathsfit}{bold}{\encodingdefault}{\sfdefault}{bx}{n}

\usepackage[framemethod=TikZ]{mdframed}

\newcommand{\bfI}{\mathbf{I}}

\newcommand{\bepsilon}{{\boldsymbol{\epsilon}}}

\makeatletter
\usepackage{xspace} 
\def\@onedot{\ifx\@let@token.\else.\null\fi\xspace}
\DeclareRobustCommand\onedot{\futurelet\@let@token\@onedot}

\usepackage{amsmath}

\newcommand{\linfty}{\mathcal{L}_{\infty}(\rvx)}

\title{
Generative Representation Learning \\ for Scalable Latent Tokenization
}

\author{
\textbf{Alan Baade}$^{1}$\thanks{Correspondence to \texttt{baade@stanford.edu}.} \quad
\textbf{Kyle Sargent}$^{1}$ \quad
\textbf{Vincent Sitzmann}$^{2}$ \quad
\textbf{Eric Ryan Chan}$^{1}$\thanks{Equal advising.} \\
\textbf{Justin Johnson}$^{3\dagger}$ \quad
\textbf{Li Fei-Fei}$^{1\dagger}$ \quad
\textbf{Gordon Wetzstein}$^{1\dagger}$ \\[4pt]
$^{1}$Stanford University \quad
$^{2}$Massachusetts Institute of Technology \\
$^{3}$University of Michigan
}

\iclrfinalcopy 
\begin{document}

\maketitle



\begin{abstract}

We introduce an approach for self-supervised representation learning designed to be used as a tokenizer for generative modeling. Our objective supervises a representation loss by factorizing the pointwise mutual information of the data, which we estimate by taking information differences between conditional predictions from a learned generative model. We justify our algorithm theoretically and empirically, demonstrating that our generative approach to representation learning avoids dimensional collapse and feature suppression in cases where commonly used methods like DINO and LeJEPA fail. Moreover, we show our representation learning technique is scalable; expending more compute leads to predictably better latent tokenization.
For the first time, we perform a careful analysis of the \emph{combined} flop efficiency of tokenizer and generative model and demonstrate that our approach is a compute-optimal improvement over direct diffusion for pixel-space generation, obtaining an FID-50K of \textbf{2.18} with JiT-L. Applied to ImageNet and the text-to-image dataset GPIC, we show tokenizer encoder scaling laws for diffusability, where extrapolating our approach achieves representations in $\mathbf{7\times}$ fewer FLOPs than DINOv2 on ImageNet and $\mathbf{8\times}$ fewer FLOPs than WebSSL on GPIC. Code at \url{https://github.com/AlanBaade/repr}
\vspace{-1em}
\end{abstract}



\section{Introduction}

Representation learning decides what information matters. Traditionally, that decision has been guided by downstream tasks: augmentation invariance, for example, encourages representations that preserve semantic content while discarding other latent variables. 
Yet representations designed this way have become a foundation for generative modeling, serving as latent tokenizers for diffusion models. Methods such as RAE \citep{rae}, for example, inherit encoders like DINOv2 \citep{oquab2024dinov2}, whose objectives were not designed to make generation easier. Moreover, these encoders can require orders of magnitude more data and compute to train than the generative models built on top of them. This leaves two basic questions unresolved: \emph{what makes a representation good for generation, and is learning one actually more compute-efficient than modeling pixels directly?
}

%


In this paper, we seek to discover whether there is a representation learning algorithm that demonstrates tokenizer encoder scaling laws: As more compute and data is given to train the tokenizer, generative models trained on these representations converge faster, and the compute optimal balance between training a tokenizer and generative model places nonzero weight on tokenization. Although prior work has shown scaling laws for tokenization \citep{vtp,vitokv2}, these works couple tokenizer decoders, which are a part of the generative process that has been widely shown to improve with scale \citep{Kaplan2020ScalingLF}, with tokenizer encoders that instead have been shown to \emph{worsen} with scale \citep{simeoni2025dinov3,vitokv2,yao2025vavae}, a phenomenon we discuss in depth. 




Toward the goal of discovering scalable encoders, we identify three primary reasons for the failures of current recipes.
\textbf{1) Feature suppression:} Standard representation learning techniques such as Contrastive Learning \citep{inproceedings,vandenOord2018RepresentationLW} and LeJEPA~\citep{balestriero2025lejepaprovablescalableselfsupervised} can have their objectives saturated by a small number of bits, behavior commonly referred to as feature suppression \citep{Geirhos2020ShortcutLI,chen2021intriguing}. Although suppressed objectives may be ideal for tasks like ImageNet probing, where only a few semantic bits need highlighting, in generative modeling all latent variables in the data are sampled.
\textbf{2) Causal Theoretical Grounding:} Representation learning algorithms like contrastive learning optimize a CLS token, but the tokenizer used by methods like RAE is instead ``emergent'' internal patch features that have demonstrated instability under scaling \citep{darcet2024vision}. 
\textbf{3) Architectural scalability:} Classic tokenizers such as VAEs \citep{kingma2022autoencodingvariationalbayes}
and VQ-VAE \citep{oord2018neuraldiscreterepresentationlearning} bottleneck gradient descent through techniques like the reparameterization trick or straight-through gradient estimation and fail to learn semantics \citep{yao2025vavae}, resulting in the transition toward tokenizers based on representation learning such as DINOv2 \citep{oquab2024dinov2}. 



To address these issues preventing scalable representations, we propose an approach to move representation learning away from discriminative tasks, such as contrastive learning, and toward generative approaches instead. 
Specifically, we observe that Word Embeddings and Contrastive Learning implicitly factorize a pointwise mutual information (PMI) matrix of the data distribution \citep{omer_neural, vandenOord2018RepresentationLW, chen2023bag}. We propose to use a generative model to directly estimate a weighted PMI that our representation learner factorizes explicitly. By factorizing the PMI of the data, we directly inherit the long-studied theoretical properties of Contrastive Learning in our latent space.
By using a generative objective instead of a discriminative objective, our approach both uses no negative samples and is robust to feature suppression, enabling our approach to directly supervise tokenizer patch feature representations rather than relying on intermediate features in a network \citep{rae} or excessively destroying information when directly using contrastive representations \citep{ramesh2022dalle}.
Our PMI estimation inherits scalability from existing generative models, we use continuous gradient descent and standard Transformer architectures, and our representation supervision loss is directly interpretable as Nats Squared.


We empirically demonstrate our approach for the task of image generation on both ImageNet \citep{deng_imagenet} and GPIC \citep{chandrasegaran2026gpicgiantpermissiveimage}, a text-to-image dataset.
We isolate our gains solely to the diffusability of the encoder by training generative models with Latent Forcing \citep{baade2026latent}, demonstrating scaling directly on pixel generation while using no decoder, perceptual losses, adversarial losses, or external models. We further show that our approach is generalizable to different settings, such as unconditional generation, guided and unguided generation, and GAN+LPIPS decoding \citep{Goodfellow2021GenerativeAN, zhang2018perceptual}. Our resulting approach provides a FLOP-wise performance improvement over JiT, obtaining an FID-50K of $2.18$ with JiT-L.

In summary, we make the following contributions:
\vspace{-0.8em}
\begin{enumerate}
    \setlength{\itemsep}{0pt}
    \setlength{\parskip}{0pt}
    \setlength{\parsep}{0pt}
    \item We introduce Relative Prediction Representations (RePR), a theoretically grounded representation learning algorithm for discovering latent variables within a signal that correlate between different masked views of the data.
    \item We provide a practical implementation on real high-dimensional image data.
    \item A careful analysis of tokenizer encoder scaling laws and the \emph{combined} FLOP efficiency of representation learning and generative modeling, improving JiT-L FID-50K on ImageNet to $2.18$ with equal FLOPs and extrapolating to equivalent tokenizer quality $7\times$ faster than DINOv2 \citep{oquab2024dinov2} on ImageNet and $8\times$ WebSSL \citep{fan2025scalinglanguagefreevisualrepresentationwebssl} on GPIC.
\end{enumerate}

\vspace{-1em}

\section{Related Work}

\textbf{Representation Learning.}
Contrastive Learners \citep{pmlr-v119-chen20j,He_2020_CVPR} are a staple of representation learning that supervise a loss objective by encouraging high similarity between augmented views of the same scene and low similarity across different scenes. Empirically improving on contrastive-style objectives has seen great success in the widely-used DINO v1,v2,v3 series of models \citep{Caron_2021_ICCV,oquab2024dinov2,simeoni2025dinov3}. A longstanding flaw of contrastive objectives is that they struggle with strongly conditioned views, where hard negative examples \citep{robinson2021contrastive} that prevent trivializing the objective are difficult to find. As a result, contrastive learning objectives often exhibit feature suppression \citep{chen2021intriguing}, where contrastive representations only capture a subset of relevant variables in a scene, for example probing well to object class but losing scene structure. To combat issues of hard negatives and feature suppression, substantial effort has been devoted to designing hand-crafted augmentations intended to highlight signal for desired variables \citep{pmlr-v119-chen20j}. Representation learning has also seen success through predictive objectives, such as predicting masked-out pixels in an image as done in Context Encoders and MAE \citep{Pathak2016ContextEF,He_2022_CVPR}, or by replacing the pixel prediction target with 
a neural network latent such as BYOL, Data2Vec, JEPA, and MAGE \citep{BYOL, data2vec,Assran_2023_CVPR, bardes2024revisiting, li2022mage}. 


\textbf{Pointwise Mutual Information and Representations}
The Pointwise Mutual Information (PMI) of the data is closely tied to many representation learning algorithms. 
The connection between factorizing the PMI and representation learning was first established by \cite{omer_neural} for Word2Vec \citep{mikolov2013efficientestimationwordrepresentationsword2vec} and afterward extended to contrastive learning \citep{vandenOord2018RepresentationLW, chen2023bag}. The Platonic Representation Hypothesis \citep{huh2024prh} proposes that large-scale neural networks emergently converge to similarly structured representations and suggests PMI factorization as a candidate for such a space. However, recent work \citep{koepke2026cave} has shown that at large scales, neural representations fail to match such structure. In this work, we directly supervise representations to match the PMI, rather than assuming emergence at scale.

\textbf{Pixel and Latent Space Generative Modeling.}
The current state-of-the-art for image generative modeling uses diffusion transformers \citep{peebles2023scalablediffusionmodelstransformers}, of which there are two main varieties: direct pixel diffusion and latent diffusion. A representative approach for pixel-space diffusion is JiT \citep{li_jit}, which demonstrated that pixel-space diffusion can be a practical and scalable method using transformer architectures. Meanwhile, latent diffusion \citep{Rombach_2022_CVPR,li_rcg} instead modifies generation into two stages, first creating a latent tokenizer, and second training a generative model on the tokenizer. Modern latent diffusion approaches like RAE \citep{rae} and REPA \citep{ICLR2025_d9e42b4d} demonstrated strong results, attaining state-of-the-art FID values on ImageNet. However, RAE and REPA do not control for representation learner pretraining, where the chosen latent space, DINOv2-b \citep{oquab2024dinov2} is distilled from a 1.1 billion parameter model trained on over 100 times more data than ImageNet. 
As a result, the FLOP-efficiency of representation encoding is not currently understood.


\vspace{-0.8em}

\section{Background}
\vspace{-0.3em}
Our method rests on the estimation of the pointwise mutual information between two views of a data point using generative models and using it to directly supervise their feature embeddings. 

\textbf{InfoNCE converges to PMI of datapoints.} The PMI is defined as the information gain on one outcome $\rvx\sim X$ when conditioned on a different outcome $\rvy\sim Y$:
\begin{align}
\operatorname{PMI}(\rvx,\rvy)
= \operatorname{PMI}(\rvy,\rvx)
= \log \frac{p_{X\mid Y}(\rvx\mid \rvy)}{p_X(\rvx)}
= \log \frac{p_{Y\mid X}(\rvy\mid \rvx)}{p_Y(\rvy)}
= \log \frac{p_{X,Y}(\rvx,\rvy)}{p_X(\rvx)\,p_Y(\rvy)}.
\end{align}
Intuitively, PMI measures how much more or less often two observations co-occur than expected if independent.
\cite{vandenOord2018RepresentationLW} showed that the dot product of contrastive embeddings $R_\theta(\rvx)$ and $R_\theta(\rvy)$ converges to an estimate of the PMI up to an unidentifiable scalar offset:
\begin{equation}
    \langle R_\theta(\rvx), R_\theta(\rvy) \rangle \approx \operatorname{PMI}(\rvx,\rvy) + a(\rvx).
    \label{eq:contrastive_PMI}
\end{equation}
\cite{omer_neural} similarly demonstrated that the dot product of word and context embedding matrices form a low-rank, weighted factorization of their PMI matrix.

\textbf{Estimating the PMI with a generative model.} Assuming that a generative model $G_\theta$ gives us access to the log probability of a data sample, $\log P_{G_\theta}(\rvy)$, this then allows us to obtain an estimate of the mutual information between two data samples $\rvx$ and $\rvy$ as follows: 
\begin{equation}
\operatorname{PMI(\rvx, \rvy)} = \log P_{G_\theta}(\rvy | \rvx) - \log P_{G_\theta}(\rvy).     
\label{eq:gen_PMI}
\vspace{-1.6em}
\end{equation}

\vspace{-0.2em}
\section{Method}
\vspace{-0.2em}
We now give a high-level overview of our algorithm.
Given a training sample, we randomly mask to obtain two corrupted views, $\rvx$ and $\rvy$. We then obtain an estimate of their PMI using a conditional generative model following Eq.~\ref{eq:gen_PMI}. In parallel, just as in contrastive learning, we embed each view into a feature using a representation encoder $R_\theta$. Combining Eqs.~\ref{eq:contrastive_PMI},~\ref{eq:gen_PMI}, we directly supervise the dot product of the embeddings $R_\theta(\rvx)$ and $R_\theta(\rvy)$ to equal the PMI estimated via the generative model:
\begin{align}
\mathcal{L}_{\mathrm{PMI}}
& =\Bigl\|\langle R_\theta(\rvx),R_\theta(\rvy)\rangle-\operatorname{PMI}(\rvx,\rvy)\Bigr\|_2^2 \\
& =\Bigl\|\langle R_\theta(\rvx),R_\theta(\rvy)\rangle-\operatorname{stopgrad}\!\Bigl(\log P_{G_\theta}(\rvy\mid \rvx)-\log P_{G_\theta}(\rvy)\Bigr)\Bigr\|_2^2.
\end{align}
Intuitively, this proposal follows the success of contrastive learning, which implicitly supervises the dot product $\langle R_\theta(\rvx),R_\theta(\rvy)\rangle$ with their PMI following Eq.~\ref{eq:contrastive_PMI}.
However, contrastive learning accomplishes this via the InfoNCE loss, which depends on the sampling of negatives, which is challenging in strongly conditional settings and thus, does not scale. We instead estimate the PMI directly with a conditional generative model.
Because the capabilities of generative models are subject to robust scaling laws~\citep{Kaplan2020ScalingLF}, we will show in our experiment section that this yields a scalable, state-of-the-art representation learning objective for tokenization.

\textbf{Overview of this section.} 
As described, the approach is not yet practical for two main reasons. First, high-dimensional modalities like images have many high-frequency bits that dominate their PMI, overshadowing bits we may care more about such as semantics. We address this in Secs.~\ref{ref:pmi_estimation} and \ref{sec:timestep_weighting}. Second, likelihood estimation is expensive or intractable on large inputs such as when a view $\rvx$ is the left half of an image or the first half of a movie. We resolve this via a novel bootstrapping method discussed in Sec.~\ref{sec:bootstrapping}. Finally, we assemble the complete objective in Sec.~\ref{sec:repr_obj}.

\subsection{PMI Estimation from Generative Models}
\label{ref:pmi_estimation}
\vspace{-0.5em}


In this section, we will briefly review how to extract probabilities from generative models. We assume that all data is discrete: images have 24-bit pixels and model latents are represented by finite-precision floating points. Conditional likelihood, and therefore PMI, is simple to estimate from discrete cross-entropy models for both text \citep{devlin-etal-2019-bert} and images \citep{li2025fractalmar} which output log-probabilities for individual tokens given a context.

We estimate probabilities from diffusion models using the ELBO, which provides exact probabilities on discrete data \citep{kingmavdm}, which we discuss below. This is different from probability-flow ODE methods, which integrate the divergence of the probability-flow vector field. \citep{song2021scorebased}, and instead allows us to estimate probabilities directly from the loss of the diffusion model. Following \cite{kingmavdm} (Eq.~37) the variational lower bound (VLB) can be expressed as: 
\begin{align}
\label{diff_vlb}
    - \log p(\rvx) \leq %
    - \text{VLB}(\rvx) =
    \underbrace{%
    D_{KL}(q(\rvz_1|\rvx)||p(\rvz_1))
    }_{\text{Prior loss}}
    + \underbrace{%
    \E_{q(\rvz_0|\rvx)}\left[ - \log p(\rvx|\rvz_0) \right]
    }_{\text{Reconstruction loss}}
    +
    \underbrace{
    \linfty.
    }_{\text{Diffusion loss}}
\end{align}
%
Moreover, the bound is tight, i.e., $\log p(\rvx) = \text{VLB}(\rvx)$, when the denoising network is optimal and the endpoint terms vanish~\citep[Eqs.~9,10]{song2021maximum}. In this work we use the linear-interpolation (flow-based) schedule $\rvz_t = (1-t)\rvx + t\bepsilon$, with $\bepsilon \sim \mathcal{N}(0, \bfI)$. When $t=1$ (at full noise), since $\rvz_1 = \bepsilon$, we have $q(\rvz_1 \mid \rvx) = p(\rvz_1)$ and the prior loss is zero. At $t=0$, since $\rvz_0 = \rvx$ and the data is discrete, $\rvx$ can be recovered exactly from $\rvz_0$, so the reconstruction loss is also zero. The VLB therefore reduces to the diffusion loss $\linfty$. Under this noise schedule, following \citep[Eqs.~53,54]{kingmavdm}, the diffusion loss expands as:
%
\begin{align}
    \log p(\rvx) &= \text{VLB}(\rvx) = -\linfty = -\E_{t\sim \mathbf{U}[0,1]\bepsilon\sim\mathcal{N}(0,\bfI)}\left[ \frac{1-t}{t^3} \left\rVert \rvx - \E(\rvx \mid \rvx_t,t) \right\lVert_{2}^{2}\right].
\end{align}
The PMI between two points can then be calculated by subtracting their log probability estimates:
\begin{align}
\label{pmidiffusion}
    \operatorname{PMI}(\rvx,\rvy) = \E_{t\sim \mathbf{U}[0,1],\bepsilon\sim\mathcal{N}(0,\bfI)}\left[ \frac{1-t}{t^3} \left( \left\rVert \rvy - \E(\rvy \mid \rvy_t;t) \right\lVert_{2}^{2} - \left\rVert \rvy - \E(\rvy \mid \rvx,\rvy_t,t) \right\lVert_{2}^{2} \right) \right].
\end{align}
%


Here, we obtain the outer expectation over $t,\bepsilon$ through Monte Carlo sampling, and the inner expectations using a learned denoising model. Although the VLB is only equal to $\log p(\rvx)$ with an optimal denoiser, our approximation approaches the true PMI as we improve the prediction.




\subsection{Timestep Weighting}
\label{sec:timestep_weighting}
\vspace{-0.5em}

Practical implementations of diffusion models do not directly optimize the ELBO but rather a timestep-weighted version \citep{Karras2022edm}. While likelihood weights every bit of information equally, information at lower signal-to-noise ratios (SNRs) often carries greater semantic or perceptual importance, and therefore, diffusion models benefit by biasing a finite-capacity model towards the most important SNRs. Likewise, the standard PMI treats all shared information equally, including fine-grained information that is only predictable at very high SNRs. 
For representation learning, we instead introduce an inductive bias for which SNRs determine similarity. We let:
\begin{align}
\mathbf{y}_s
&= (1-s)\mathbf{y} + s\boldsymbol{\epsilon},
\qquad
\boldsymbol{\epsilon} \sim \mathcal{N}(0,I).
\end{align}
For each corruption level $s$, we can define
\begin{align}
\operatorname{PMI}(\mathbf{x},\mathbf{y}_s)
&=
\log
\frac{
p_s(\mathbf{y}_s \mid \mathbf{x})
}{
p_s(\mathbf{y}_s)
}.
\end{align}
At $s=0$, this reduces to the ordinary PMI. As $s$ increases, fine-grained information is progressively destroyed, so the PMI measures only dependencies that remain predictive under stronger corruption. We therefore define a scale-weighted PMI,
\begin{align}
\operatorname{PMI}^{*}(\mathbf{x},\mathbf{y})
&=
\mathbb{E}_{s\sim U[0,1],\,\boldsymbol{\epsilon}}
\left[
q(s)\operatorname{PMI}(\mathbf{x},\mathbf{y}_s)
\right],
\end{align}
where $q(s)$ determines which information scales are emphasized. Each $\operatorname{PMI}(\mathbf{x},\mathbf{y}_s)$ can itself be estimated using the diffusion ELBO, which integrates the conditional denoising advantage over a second diffusion timestep $t$. 
More generally, because the corruption at $s$ and the diffusion corruption at $t$ are both Gaussian, they can be composed into a single effective timestep weighting: 
\begin{align}
\operatorname{PMI}^{*}(\mathbf{x},\mathbf{y})
&=
\mathbb{E}_{t\sim f(t),\,\boldsymbol{\epsilon}}
\left[
g(t)
\left(
\left\|
\mathbf{y}
-
\mathbb{E}(\rvy\mid\mathbf{y}_t,t)
\right\|_2^2
-
\left\|
\mathbf{y}
-
\mathbb{E}(\rvy\mid\mathbf{y}_t,t,\mathbf{x})
\right\|_2^2
\right)
\right].
\label{eq:timestepweighting}
\end{align}
where $f(t)$ is timestep sampling and $g(t)$ timestep weighting, providing a practical way to bias the representation objective toward information that remains shared across the noise scales of interest.




\subsection{Bootstrapped PMI Estimation}
\label{sec:bootstrapping}
\vspace{-0.5em}
Likelihood estimation gets harder as the amount of information in the input grows: predicting the probability of a single pixel is easy and calculable with dataset statistics, but predicting the likelihood of an image using a diffusion model may take intractably many Monte Carlo Estimation steps. To avoid this issue, we replace a challenging full-view PMI estimate with a sequence of local PMI increments. Instead of estimating the PMI between $\rvx$ and an entire large target $\rvy$, we write $\rvy$ as tokens $y_1, ..., y_k$ and supervise only the \emph{incremental} PMI contributed by the next token: 
%
\begin{flalign}
&\left\langle R_\theta(\rvx),
R_\theta(y{\le k})-R_\theta(y_{\le k-1})\right\rangle
=
\operatorname{PMI}(\rvx,y_{\le k})
-\operatorname{PMI}(\rvx,y_{\le k-1})
&&\notag\\
&\qquad=
\sum_{i=1}^{k}\log\frac{p(y_i\mid y_{<i},\rvx)}{p(y_i\mid y_{<i})}
-\sum_{i=1}^{k-1}\log\frac{p(y_i\mid y_{<i},\rvx)}{p(y_i\mid y_{<i})}
=
\log\frac{p(y_k\mid y_{<k},\rvx)}{p(y_k\mid y_{<k})}
=
\operatorname{PMI}(y_k,\rvx\mid y_{<k}).
&&
\end{flalign}
Because these increments telescope, summing them recovers the full PMI:
\begin{equation}
\sum_{k=1}^K
\left[
\operatorname{PMI}(\rvx,y_{\le k})
-
\operatorname{PMI}(\rvx,y_{\le k-1})
\right]
=
\operatorname{PMI}(\rvx,y_{\le K})
-
\operatorname{PMI}(\rvx,\varnothing)
=
\operatorname{PMI}(\rvx,\rvy).
\end{equation}
This makes the likelihood-ratio target easier to estimate than the PMI of the whole input at once. 

\subsection{Representation Objective} 
\label{sec:repr_obj}

We combine Timestep Weighting and Bootstrapping to obtain our full representation objective. Given input views $\rvx$ and $\rvy$ and an incremental amount of information to add $\rvz$, we optimize for:
%
\begin{equation}
\label{eq:repr_loss}
\begin{aligned}
\mathcal{L}_{\mathrm{RePR}}
&= \Bigl\|
    \left\langle R_\theta(\rvx),
    R_\theta(\rvy,\rvz)-R_\theta(\rvy)\right\rangle
    - \operatorname{PMI}^*(\rvx, \rvz \mid \rvy)
    \Bigr\|_2^2 \\
&= \Bigl\|
    \left\langle R_\theta(\rvx),
    R_\theta(\rvy,\rvz)-R_\theta(\rvy)\right\rangle \\
& \>\>\>\>- \mathbb{E}_{t\sim f(t),\,\boldsymbol{\epsilon}}
    \Bigl[
    g(t)\operatorname{stopgrad}\Bigl(
    \left\|\mathbf{z}
    - \mathbb{E}(\rvz\mid\mathbf{z}_t,t,\rvy)\right\|_2^2
    - \left\|\mathbf{z}
    - \mathbb{E}(\rvz\mid\mathbf{z}_t,t,\mathbf{x},\rvy)\right\|_2^2
    \Bigr)
    \Bigr]
    \Bigr\|_2^2.
\end{aligned}
\end{equation}

\vspace{-1.5em}
\section{Architecture}
\label{sec:architecture}

\begin{figure}[t]
    \centering
    \includegraphics[width=0.98\textwidth, trim={0.0cm 3.7cm 0.0cm 4.6cm}, clip]{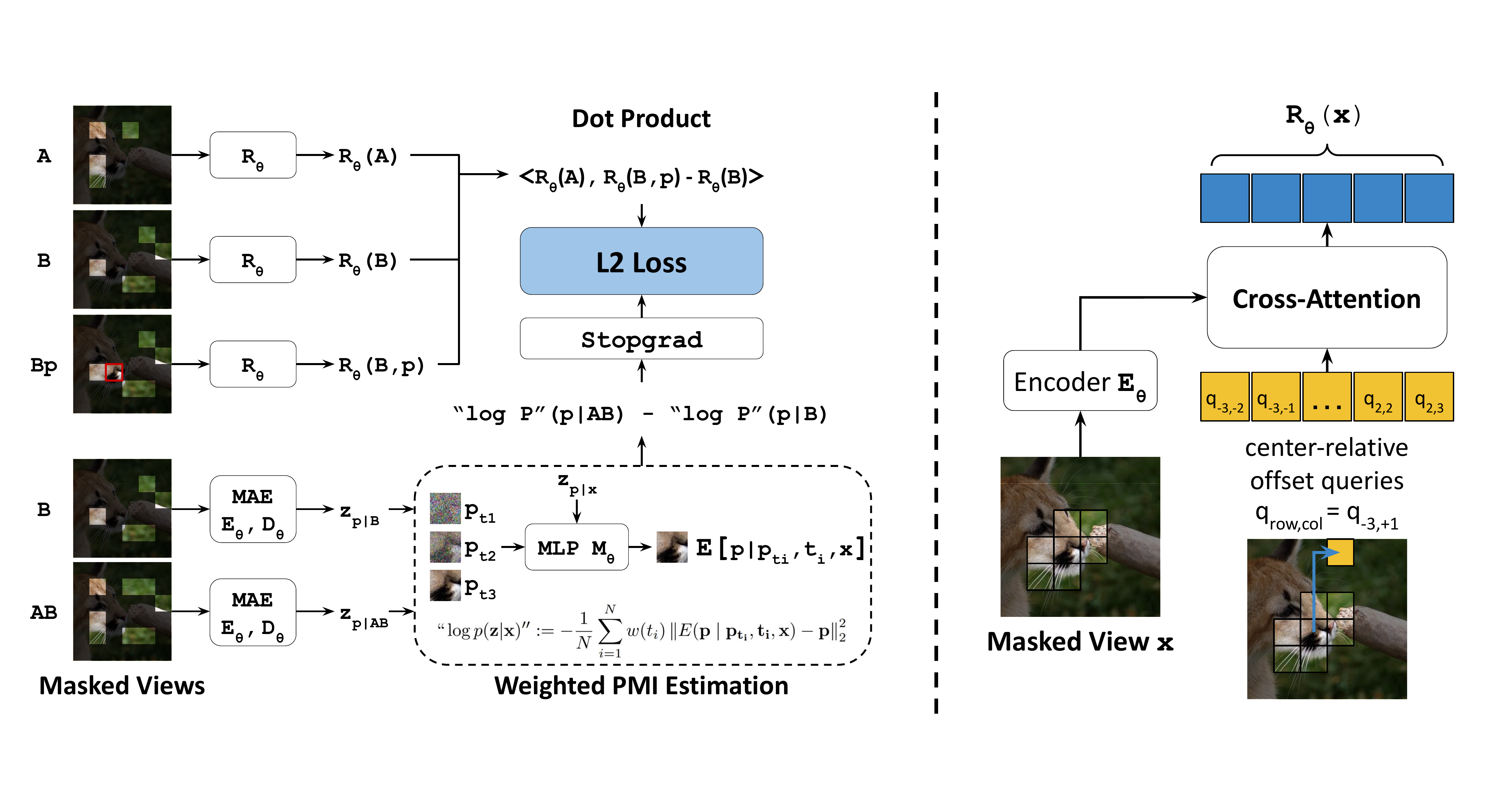}
    \caption{Architecture. \textbf{Left:} 1) Our Prediction Model (bottom) estimates weighted data likelihoods for a target patch $p$ using an MAR architecture, which combines an MAE Encoder $E_\theta$ and Decoder $D_\theta$ with a diffusion prediction MLP head $M_\theta$. 2) The Representation Model $R_\theta$ (top) outputs vector representations supervised to factorize a weighted PMI. \textbf{Right:} We enforce spatial equivariance in $R_\theta$ by hiding absolute position information but outputting a feature for every patch.}
    \label{fig:architecture}
    \vspace{-1em}
\end{figure}



Here, we describe how we instantiate RePR for learning representations from images. RePR consists of two learned networks: a ``Prediction Model" that predicts masked inputs given partial observations (to estimate the PMI), and a ``Representation Model'' that outputs vectors given a partial input (into which we distill the PMI).  In the following sections, we describe the Prediction Model (Sec.~\ref{sec:prediction_model}), the Representation Model for patch features (Sec.~\ref{sec:patch_features}), how we mask views for PMI calculation (Sec.~\ref{sec:masking_strategy}), and estimating the PMI with respect to latents for improved performance (Sec.~\ref{sec:feature_prediction}).

\vspace{-0.5em}
\subsection{Prediction Model}
\label{sec:prediction_model}
\vspace{-0.5em}

%

The ``Prediction Model'' is implemented as a MAR~\citep{li_mar}, a modification of Masked Autoencoders \citep{He_2022_CVPR} that replaces the patch regression head with a patch diffusion MLP. Specifically, MAR consists of an MAE Encoder $E_\theta$ (without mask tokens), MAE Decoder $D_\theta$ (with mask tokens) and a diffusion MLP $M_\theta$. The diffusion MLP takes in the decoder latent corresponding to patch $p, z_p$, a noised view of $p$, $p_t$, and trains to denoise $p$, where we follow JiT~\citep{li_jit} and use $x$-prediction with a v-loss, depicted in Figure \ref{fig:architecture}.
\begin{equation}
\label{eq:mar_loss}
    \mathcal{L}_\text{MAR} = w(t)||M_\theta(z_p,p_t,t) - p||_2^2.
\end{equation}
MAR is convenient for estimating PMI because when estimating likelihoods for a contextualized latent, the cost of the Encoder and Decoder is shared, after which we can run the lightweight diffusion loss head many times to average across different noise levels.

\subsection{Representing Patch Features}
\label{sec:patch_features}
\vspace{-0.5em}

We distinguish between features that are directly supervised for structure, such as the CLS token of Contrastive Learners, DINO, or LeJEPA, and unregularized intermediate latents, such as the non-CLS tokens of DINO or the latents in an unregularized autoencoder. Without explicit structural supervision, these unregularized latents can diverge or collapse. For example, DINOv2 demonstrated outlier patch features, motivating the introduction of register tokens \citep{darcet2024vision}, and in $\text{DINOv3}$ \citep{simeoni2025dinov3}, feature quality again degraded, necessitating an additional distillation stage to align model features with an earlier checkpoint, taken before degradation occurred. While heuristic regularization and interventions can correct divergence, these issues may become increasingly difficult to control with greater scale. Such feature instability motivates directly supervising patch features for principled and scalable tokenization.

We directly use the full patch features $R^{H\times W \times D}$ from our network as our representation vector. To enforce spatial structure, we implement an additional spatial equivariance constraint: We remove absolute position information from the Encoder $E_\theta$, however we take the dot product of patch features at their absolute positions. To implement this, our Representation Model cross-attends from output of the encoder $E$, to learned position embeddings indexed by their relative offset to the center of the visible input, visualized in Fig.~\ref{fig:architecture}, right.

\vspace{-0.6em}
\subsection{Masking Strategy} 
\label{sec:masking_strategy}
\vspace{-0.5em}

For a given input image $X$, during masking we produce two potentially overlapping patch masks $A$ and $B$, as well as a non-overlapping target patch $p$ that is denoised by the MAR head. Following MAR, we use a randomized masking strategy
where the masking ratio is randomly sampled from $U[M_A,100\%]$ and $U[M_B,100\%]$ respectively. By randomly choosing the masking ratio instead of fixing a single masking ratio, we supervise a range of context sizes to reconstruct the telescoping series in Sec.~\ref{sec:bootstrapping}. For additional efficiency, we observe that we can reuse the calculation for $E(Bp)$ for several different views $A_i$, and we can reuse the decoder pass to predict several target patches $p_j$ for only additional integration cost. In our final implementation we therefore add hyperparameters $A_\text{Repeat}$ and $p_\text{Repeat}$ for efficiency. Algorithm code in Appendix~\ref{alg:repr}


\vspace{-0.7em}
\subsection{Iterative Feature Prediction} 
\label{sec:feature_prediction}
\vspace{-0.5em}

We find that recursively applying our training technique, by replacing the predicted incremental patches with features from a previous cycle of training, can yield additional gains. This mirrors findings in representation learning, where self-supervision by predicting in a feature space \citep{hubert,zhou2021ibot,data2vec} has been shown to strongly improve representation quality over direct pixel prediction objectives such as MAE \citep{He_2022_CVPR}.
Because a factorization of the PMI can reconstruct the input at optimality (App.~\ref{app:pmi_properties}), unweighted PMI estimation for latent prediction is the same as for pixel prediction: $\operatorname{PMI}(\rvx,R_\theta(\rvy))=\operatorname{PMI}(\rvx,\rvy)$. With timestep weighting and bootstrapped prediction, feature-space prediction provides further control over learning signal.
In practice, we embed the entire image $R_{\theta_\text{old}}(\rvx)$ and substitute incremental patch $p$ for the patch representation corresponding to the same location in the image. By sharing feature calculation between $A_\text{Repeat}$ views, embedding with $R_{\theta_\text{old}}$ is implementationally cheap.
\vspace{-0.7em}
\subsection{Joint Optimization}
\vspace{-0.5em}




Our approach trains the Prediction Model and the Representation Model jointly, simultaneously optimizing the losses in Eq.~\ref{eq:mar_loss} and \ref{eq:repr_loss}. Empirically, we find the Representation Model converges relatively easily, therefore to speed training, we lazily optimize, training the Representation Model $R_\theta$ only once every four training iterations.

\vspace{-0.8em}
\section{Experiments}
We empirically evaluate our approach for the task of image generation on ImageNet \citep{deng_imagenet} and the text-to-image dataset GPIC \citep{chandrasegaran2026gpicgiantpermissiveimage}. In this section, we demonstrate that RePR is robust to Feature Suppression (Sec.~\ref{sec:feature_suppression}), provide scaling laws for RePR as a Tokenizer Encoder (Sec.~\ref{sec:scaling_laws}), and use RePR to obtain compute-optimal pixel diffusion (Sec.~\ref{sec:compute_optimal}).

\vspace{-0.5em}
\subsection{Feature Suppression}
\vspace{-0.5em}
\label{sec:feature_suppression}
\begin{wrapfigure}{r}{0.50\linewidth}
    \vspace{-2.4em}
    \centering
    \includegraphics[width=\linewidth, trim={0.75cm 0.5cm 1cm 0.7cm}, clip]{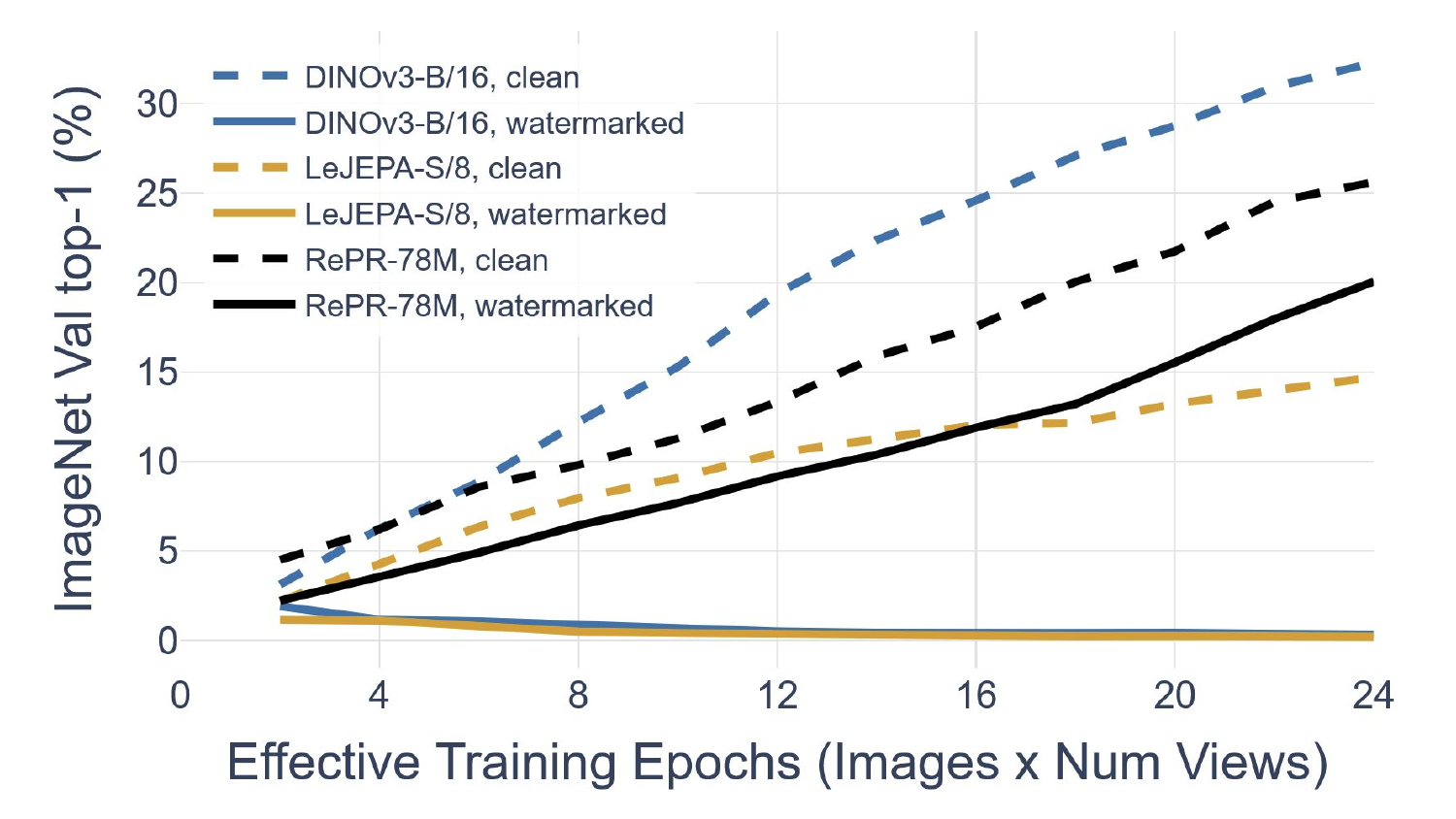}
    \caption{Discriminative representation learning can suffer from feature suppression when a small subset of features, such as a watermark, is sufficient to distinguish examples. This degrades representation quality, as reflected by collapsing ImageNet class-probing performance. RePR remains robust to feature suppression.}
    \label{fig:feat_suppression}
    \vspace{-1.2em}
\end{wrapfigure}

Existing representation learning methods such as contrastive learning suffer from an issue called Feature Suppression \citep{Geirhos2020ShortcutLI,chen2021intriguing,li_featuresuppression}  where a self-supervised objective might learn only a minimal subset of features necessary to solve the downstream task (e.g. instance discrimination), while dropping the remaining latent variables. This can be disastrous when training tokenizers because the additional latent variables, while perhaps unnecessary for satisfying the contrastive objective, may be essential for reconstructing the data.

To show our method is robust to Feature Suppression, we benchmark on a version of Imagenet that adds a visual watermark to every image patch that uniquely identifies the image in the dataset~\citep{enigma2026obsessedencoder}. On such a dataset, contrastive losses become trivialized: merely identifying the image patch watermark is enough to attain a perfect loss. In Fig.~\ref{fig:feat_suppression}, we train LeJEPA \citep{balestriero2025lejepaprovablescalableselfsupervised}, DINOv3 \citep{simeoni2025dinov3}, and our proposed RePR approach on both standard ImageNet and on the watermarked version. On watermarked data, LeJEPA and DINOv3 collapse to only representing the shared watermark. Meanwhile, despite sharing the same PMI factorization guarantees as contrastive learners, our proposed RePR is robust. 
\subsection{Scaling Laws}
\label{sec:scaling_laws}
\vspace{-0.5em}
\begin{figure}[h]
    \centering
    \includegraphics[width=0.98\textwidth, trim={0.8cm 2.8cm 0.8cm 2.8cm}, clip]{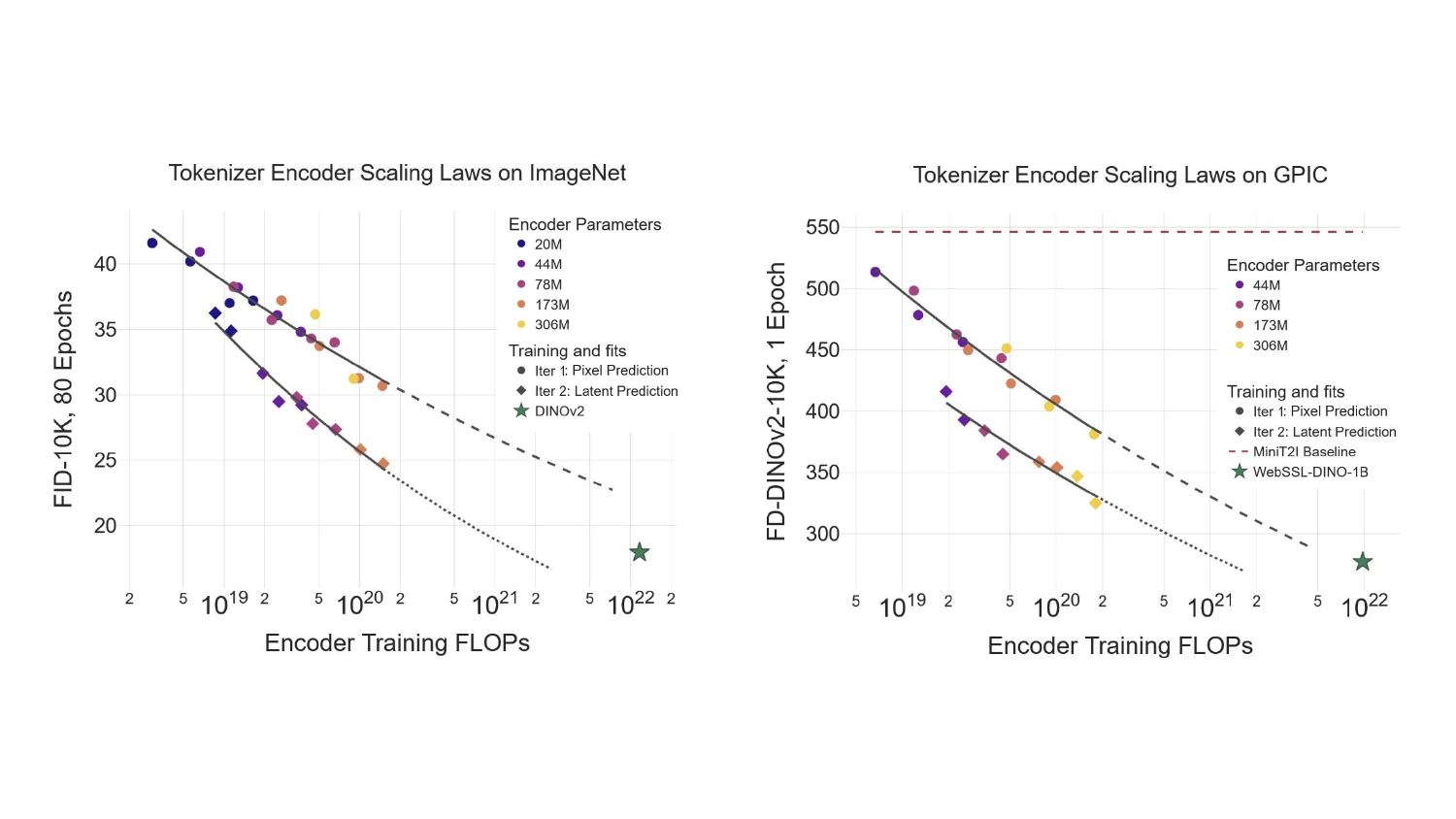}
    \caption{\textbf{Tokenizer Encoder Scaling Laws:} 
    RePR predictably scales as a tokenizer encoder. When holding size and training duration of a generative model constant, greater RePR size and FLOPs improve generation performance. Iterative feature prediction (Sec.~\ref{sec:feature_prediction}) yields further improvement.
    }
    \label{fig:scaling_laws}
    \vspace{-1.5em}
\end{figure}

In the previous section, we demonstrated that our method resists feature suppression on toy data, where prior representations collapse. In this section, we demonstrate encoder scaling laws: our model captures enough data for generation and predictably improves with scale on real data.




We demonstrate encoder scaling laws using Latent Forcing \citep{baade2026latent}, which first denoises in latent space and then generates pixels conditioned on the denoised latent. This lets us leverage the improved diffusability of learned representations while retaining generation directly in pixel space. The resulting generative process is lossless, avoiding the reconstruction--generation tradeoff of lossy latent tokenizers \citep{yao2025vavae} and isolating gains from diffusability rather than compression. Latent Forcing additionally enables controlled FLOP- and parameter-matched comparisons with direct pixel diffusion models such as JiT.

We train RePR encoders from scratch across a range of compute scales. Following $\mu$P \citep{yang2022tensorprogramsvtuning}, we perform width-only scaling, sweeping widths from 256 to 1024 while holding depth fixed. At maximum width, the inference architecture matches a standard ViT-L. Full architectural and training details are provided in Appendix~\ref{app:implementation_details}.

Fig.~\ref{fig:scaling_laws} shows consistent power-law scaling across roughly two orders of magnitude of training compute on both ImageNet and GPIC. On ImageNet, we evaluate diffusability by training a JiT-Base diffusion transformer for 80 epochs; on GPIC, we use the MiniT2i architecture \citep{minit2i2026} and train for one epoch. At each width, we additionally use feature prediction (Sec.~\ref{sec:feature_prediction}) to predict features from a pixel model of the same size trained for 200 epochs, yielding a substantial improvement over the first iteration. Extrapolating the observed trend by at most $1\times$ beyond our measured log-scale FLOP range predicts DINOv2-level performance at approximately one-seventh the training FLOPs and WebSSL-level performance at roughly one-eighth. On ImageNet, the width-1024 model begins to overfit after 200 epochs, consistent with prior scaling-law observations \citep{kim2026pretraining}; scaling continues beyond this regime on GPIC.




\vspace{-0.5em}
\subsection{Compute Optimal Generation}
\label{sec:compute_optimal}

\begin{wrapfigure}{R}{0.50\linewidth}
    \vspace{-3.5em}
    \centering
    \includegraphics[width=\linewidth, trim={7.3cm 3.2cm 7.3cm 3.2cm}, clip]{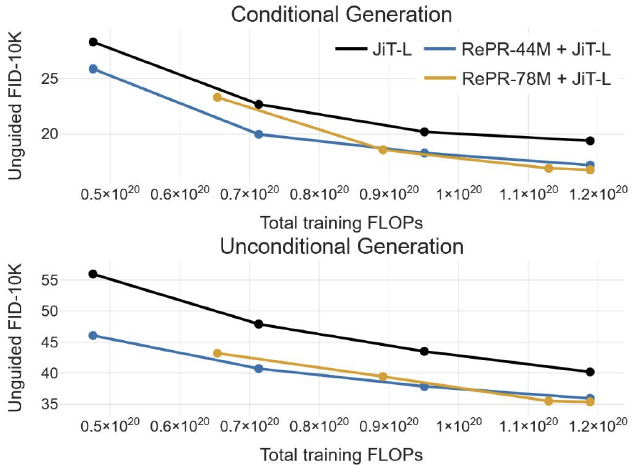}
    \caption{Total (Encoder + Diffusion) training compute versus FID-10K for JiT and JiT with Latent Forcing using RePR, showing that representation learning is compute-optimal for generation.}
    \label{fig:compute_vs_jit}
    \vspace{-2.em}
\end{wrapfigure}





To the best of our knowledge, we provide the first demonstration that representation learning combined with generative modeling can be more FLOP-efficient than direct generative modeling, even after accounting for the compute required to train both the representation and generative model. 
We isolate the comparison by using Latent Forcing~\citep{baade2026latent}, without perceptual losses \citep{Johnson2016PerceptualLF}, GANs \citep{Goodfellow2021GenerativeAN}, or external decoders.
In Fig.~\ref{fig:compute_vs_jit}, we compare FID-10K on ImageNet between JiT-Large and Latent Forcing with two RePR tokenizer sizes for conditional and unconditional generation at equal FLOPs. At a compute budget of $1.19\times 10^{20}$ FLOPs, equal to 200 epochs of JiT-L training, RePR with JiT-L improves unguided FID-10K by 2.61 for conditional generation and 4.18 for unconditional generation versus JiT-L on pixels. In Tab.~\ref{tab:jitbaseline}, we demonstrate that our approach outperforms raw pixel diffusion with guidance, attaining an FID-50k of 2.18. Together with the encoder scaling laws established above, these results identify representation learning as an additional scaling axis that can improve generative performance more efficiently than scaling the generative model alone.


\vspace{-0.7em}
\subsection{Ablation Studies}
\vspace{-0.5em}
\newcommand{\AblationMainCaptionStyle}{\captionsetup{font=normalsize,skip=4pt}}
\newcommand{\AblationMaxHeight}{0.3\textheight}
\newcommand{\AblationFont}{\scriptsize}
\newcommand{\AblationRowGap}{5pt}

\newlength{\AblationWidth}
\setlength{\AblationWidth}{1.0\textwidth}

\newsavebox{\AblationCaptionBox}
\newlength{\AblationBodyHeight}

\begin{table*}[t]
\centering

\sbox{\AblationCaptionBox}{%
\begin{minipage}{\AblationWidth}
\AblationMainCaptionStyle
\captionsetup{position=top}
\caption{\textbf{Ablation study for RePR Training.} Default configuration in \colorbox[gray]{.9}{gray}, best results \textbf{bolded}. Ablations (a,b,c) train RePR-78M for 100 epochs and (d,f) for 200 epochs. Due to the cost of training generative models, we use ImageNet \citep{deng_imagenet} Top-1 Accuracy as a proxy metric.}
\label{tab:ablation}
\end{minipage}%
}

\setlength{\AblationBodyHeight}{%
    \dimexpr\AblationMaxHeight
    -\ht\AblationCaptionBox
    -\dp\AblationCaptionBox\relax
}

\usebox{\AblationCaptionBox}\par\nointerlineskip
\begin{adjustbox}{max totalsize={\AblationWidth}{\AblationBodyHeight}}
\begin{minipage}{\AblationWidth}
\AblationFont
\setlength{\tabcolsep}{2pt}
\setlength{\aboverulesep}{1.5pt}
\setlength{\belowrulesep}{1.5pt}
\renewcommand{\arraystretch}{0.95}
\captionsetup[subtable]{
    font=ablation,
    justification=raggedright,
    singlelinecheck=false,
    skip=2pt
}

\begin{subtable}[t]{0.32\linewidth}
\centering
\begin{tabular}[t]{@{}cc@{}}
\toprule
Masking $M_A$ & IN Top-1 Acc \\
\midrule
62.5\% & 58.62 \\
75\%   & \cellcolor[gray]{.9}\textbf{58.75} \\
87.5\% & 56.21 \\
\bottomrule
\end{tabular}
\caption{\textbf{Masking for view $A$.} Typical masking ratios for MAEs \citep{He_2022_CVPR} work best}
\label{tab:placeholder-a}
\end{subtable}\hfill
\begin{subtable}[t]{0.32\linewidth}
\centering
\begin{tabular}[t]{@{}cc@{}}
\toprule
Masking $M_B$ & IN Top-1 Acc \\
\midrule
75\%    & 56.78 \\
93.75\% & \cellcolor[gray]{.9}\textbf{58.75} \\
98.5\%  & 58.55 \\
\bottomrule
\end{tabular}
\caption{\textbf{Masking for view $B$.} Higher masking rates $M_B$ perform better.}
\label{tab:placeholder-b}
\end{subtable}\hfill
\begin{subtable}[t]{0.32\linewidth}
\centering
\begin{tabular}[t]{@{}lc@{}}
\toprule
Mask strategy & IN Top-1 Acc \\
\midrule
Uniform random & 57.63 \\
Block \citep{bao2022beit}
    & \cellcolor[gray]{.9}\textbf{58.75} \\
\bottomrule
\end{tabular}
\caption{\textbf{Masking Strategies.} Block masking improves performance compared to uniform random masking.}
\label{tab:placeholder-c}
\end{subtable}

\par\vspace{\AblationRowGap}

\begin{subtable}[t]{0.27\linewidth}
\centering
\begin{tabular}[t]{@{}cc@{}}
\toprule
RePR loss weight & IN Top-1 Acc  \\
\midrule
0.025 & 63.0 \\
0.005 & \cellcolor[gray]{.9}63.1 \\
0.01  & \textbf{63.6} \\
0.1   & 49.5 \\
0.4   & 43.9 \\
\bottomrule
\end{tabular}
\caption{\textbf{RePR loss weight.} A small RePR loss weight performs better. 0.4 balances loss gradient magnitude.}
\label{tab:placeholder-d}
\end{subtable}\hfill
\begin{subtable}[t]{0.36\linewidth}
\centering
\begin{tabular}[t]{@{}llcc@{}}
\toprule
Iter. & Target & Epochs & FID-10K \\
\midrule
1 & Pixel & $\{600\}$ & 30.68 \\
2 & Iter 1 Latent & $\{200,400\}$ & 24.73 \\
3 & Iter 2 Latent & $\{200,200,200\}$ & \cellcolor[gray]{.9}\textbf{23.81} \\
\bottomrule
\end{tabular}
\caption{\textbf{Additional Training Iterations.} Calculating PMI in a feature space is more performant. Each iteration predicts the previous iteration's features at 200ep, holding total epochs constant. RePR-173M.}
\label{tab:additers-e}
\end{subtable}\hfill
\begin{subtable}[t]{0.33\linewidth}
\centering
\begin{tabular}[t]{@{}llc@{}}
\toprule
Eq.~\ref{eq:timestepweighting} $f(t)$ & Eq.~\ref{eq:timestepweighting} $g(t)$  & FID-10K \\
\midrule
$U[0,1]$ & $(1$-$t)/t^3$ & NaN \\
$U[0,1]$ & $(1$-$t)/\text{max}($0.05$,t)^3$ & 42.15 \\
$\mu$=0.6,$\sigma$=1.2 & $(1$-$t)/\text{max}($0.33$,t)^2$ & \textbf{35.09} \\  
$\mu$=1.2,$\sigma$=1.2 & $(1$-$t)/\text{max}($0.33$,t)^2$
    & \cellcolor[gray]{.9}{35.72} \\
\bottomrule
\end{tabular}
\caption{\textbf{Timestep Weighing.} Following a PMI-like timestep weighting collapses. Biasing the timestep schedule improves performance.}
\label{tab:placeholder-f}
\end{subtable}

\end{minipage}
\end{adjustbox}
\end{table*}

We evaluate key decisions in our approach in Table \ref{tab:ablation}. Variations on the masking strategy, such as the maximum visible number of patches for $A$ and $B$, are empirically useful but not critical. The loss weight for the representation loss is optimal at quite small values, suggesting that the difficulty of learning the objective largely lies in generation and not learning latent factorizations. However, representations strongly help the generative component as we see in Tab~\ref{tab:additers-e}, where training the model for a second iteration and third iteration greatly improves performance, mirroring the gains from pixel prediction approaches like MAE \cite{He_2022_CVPR} to latent-predictive approaches like Data2Vec \citep{data2vec}. 



\subsection{GAN+LPIPS Decoding} 
\vspace{-0.4em}



\newcommand{\TablePairWidth}{\textwidth}
\begin{table*}[t]
\centering
\footnotesize
\setlength{\tabcolsep}{3pt}
\renewcommand{\arraystretch}{1}
\captionsetup{font=footnotesize,position=top,skip=4pt}
\begin{minipage}{\TablePairWidth}
\begin{minipage}[t]{0.44\linewidth}
\vspace{0pt}
\caption{RAE comparisons at 20ep of DiT-XL training for \textbf{ImageNet-only encoders}. IJEPA, MocoV3 results from RAEv2 \citep{singh2026raev2}}
\label{tab:rae}
\centering
\begin{adjustbox}{max width=\linewidth}
\begin{tabular}{@{}lr@{}}
\toprule
\textbf{RAE Encoder} & \textbf{FID-50k} \\
\midrule
Mocov3-B \citep{chen2021empiricalstudytrainingselfsupervised} & 13.84 \\
IJEPA-H \citep{Assran_2023_CVPR} & 12.46 \\
RePR-173M Iter 3 (Ours) & \textbf{11.13} \\
\bottomrule
\end{tabular}
\end{adjustbox}
\end{minipage}\hfill
\begin{minipage}[t]{0.54\linewidth}
\vspace{0pt}
\caption{Representation learning is compute optimal. Under a total FLOPs budget, RePR training with Latent Forcing \citep{baade2026latent} outperforms direct pixel diffusion using JiT \citep{li_jit} with guidance, 50K generations.}
\label{tab:jitbaseline}
\centering
\begin{adjustbox}{max width=\linewidth}
\begin{tabular}{@{}llrrr@{}}
\toprule
\textbf{Representation} & \textbf{Gen. Model} &
\textbf{FLOPs} $\downarrow$ & \textbf{FID} $\downarrow$ &
\textbf{IS} $\uparrow$ \\
\midrule
Pixel & JiT-L 600ep & $3.57\times10^{20}$ & 2.36 & 298.5 \\
RePR-78M + Pixel & JiT-L 440ep & $\mathbf{3.49\times10^{20}}$ &
\textbf{2.18} & \textbf{338.4} \\
\bottomrule
\end{tabular}
\end{adjustbox}
\end{minipage}
\end{minipage}
\vspace{-0.8em}
\end{table*}

In Table \ref{tab:rae}, we compare RePR as a tokenizer for RAE \citep{rae} with a GAN+LPIPS decoder following the ablation setup in \cite{singh2026raev2}, comparing versus the reported RAE baselines trained solely on ImageNet. In addition to our unique theoretical benefits for scaling, our approach is an empirically competitive tokenizer at small scales, outperforming 
JEPA-H, a latent predictive algorithm, and Mocov3, a contrastive learner. We note that the RAE decoder relies on two external models, LPIPS for a perceptual loss, and a DINOv1/8 pretrained GAN that uses $4.7\times10^{20}$ FLOPs, more than any model we train.





\vspace{-0.5em}
\section{Conclusion}
\vspace{-0.3em}



We introduce RePR, a theoretically motivated representation learning method for generative model tokenizers. RePR shares the theoretical guarantees of contrastive learning without requiring negative samples, while remaining robust to feature suppression in settings where prior representation learning methods collapse. We establish empirical scaling laws for diffusability and show that RePR-based generative models outperform direct diffusion on raw data at matched training FLOPs. Together, these results show that representation learning with strong theoretical foundations can coexist with and improve the scaling of generative modeling.

\section*{Acknowledgments}
This material is based upon work supported by the National Science Foundation Graduate Research Fellowship under Grant No. DGE-2146755. Compute for this project was provided by Rhoda AI. We would like to thank Yilun Du for helpful discussion on the paper.

\section*{AI use statement}
In this work, we used generative AI tools for implementing methods, particularly as a coding assistant. LLM-generated code was extensively verified by authors and primarily used for data visualization and boilerplate. The core RePR algorithm is human written. We additionally used generative AI for assisting in the verification of human-written proofs and to design or provide feedback on paper readability. Additionally, we used generative AI tools for internet search. We have reviewed all AI-assisted work. We take responsibility for the final content of this work.

\bibliography{iclr2027_conference}
\bibliographystyle{iclr2027_conference}

\newpage
\appendix
\section{Appendix}






\subsection{Algorithm}

\definecolor{codeblue}{rgb}{0.25,0.5,0.5}
\definecolor{codesign}{RGB}{0,0,255}
\definecolor{codefunc}{rgb}{0.85,0.18,0.50}

\newcommand{\codefontsize}{8}
\newcommand{\codelinespacing}{8.8}

\lstdefinelanguage{RePRPython}{
  language=Python,
  keywordstyle=\color{blue}\bfseries,
  commentstyle=\color{codeblue},
  stringstyle=\color{orange},
  showstringspaces=false,
  morekeywords=[2]{get_masks,E,D,R,M,sample_t_e,sample_t_eps,mean,w_m,no_grad,repeat,w,dot},
  keywordstyle=[2]\color{codefunc},
  literate=
    {*}{{{\color{codesign}*}}}{1}
    {-}{{{\color{codesign}-}}}{1}
    {+}{{{\color{codesign}+}}}{1}
}






\begin{figure}[h]
\centering
\begin{minipage}{0.57\linewidth}
\centering
\small
\begin{algorithm}[H]
\caption{RePR: Training Pseudocode\\
Minimal Implementation, see Sec.~\ref{sec:architecture} for details. $m_{AB}$: mask for $AB$. $p_i$: index of $p$. $N$: Number of Monte Carlo integration steps. Efficiency-only parameters $A_\text{Repeat}$ and $p_\text{Repeat}$ omitted for brevity. For feature prediction, the MAR target is replaced with $R_{\theta_{\text{old}}}(E_{\theta_{\text{old}}}(x))$ and $p$ is replaced with $R_{\theta_{\text{old}}}(E_{\theta_{\text{old}}}(x))[p_i]$}
\label{alg:repr}
\begin{lstlisting}[
  language=RePRPython,
  backgroundcolor=\color{white},
  basicstyle=\fontsize{\codefontsize pt}{\codelinespacing pt}\ttfamily\selectfont,
  columns=fullflexible,
  keepspaces=true,
  breaklines=true,
  captionpos=b
]
A, B, AB, p, m_AB, p_i = get_masks(X)
a, b, ab, bp = E(A), E(B), E(AB), E(B, p)
d_ab, d_b = D(ab), D(b)

x = X[m_AB]
t, eps = sample_t_eps(x)
x_t = (1 - t) * x + t * eps
x_pred = M(x_t, t, d_ab[m_AB])
mar_loss = mean(w(t) * (x_pred - x)**2)

with no_grad():
  t, eps = sample_t_eps(p, N)
  p_t = (1 - t) * repeat(p, N) + t * eps
  p_ab = M(p_t, t, d_ab[p_i])
  p_b = M(p_t, t, d_b[p_i])
  pmi = mean(g(t)*((p_b-p)**2-(p_ab-p)**2))

pred = dot(R(a), R(bp) - R(b))
repr_loss = mean((pred - pmi)**2)

loss = mar_loss + loss_weight * repr_loss
\end{lstlisting}
\end{algorithm}
\end{minipage}
\end{figure}

\subsection{Additional Experiments}

\begin{figure}[h]
    \centering
    \includegraphics[width=0.75\textwidth, trim={0.0cm 0cm 0.0cm 2.5cm}, clip]{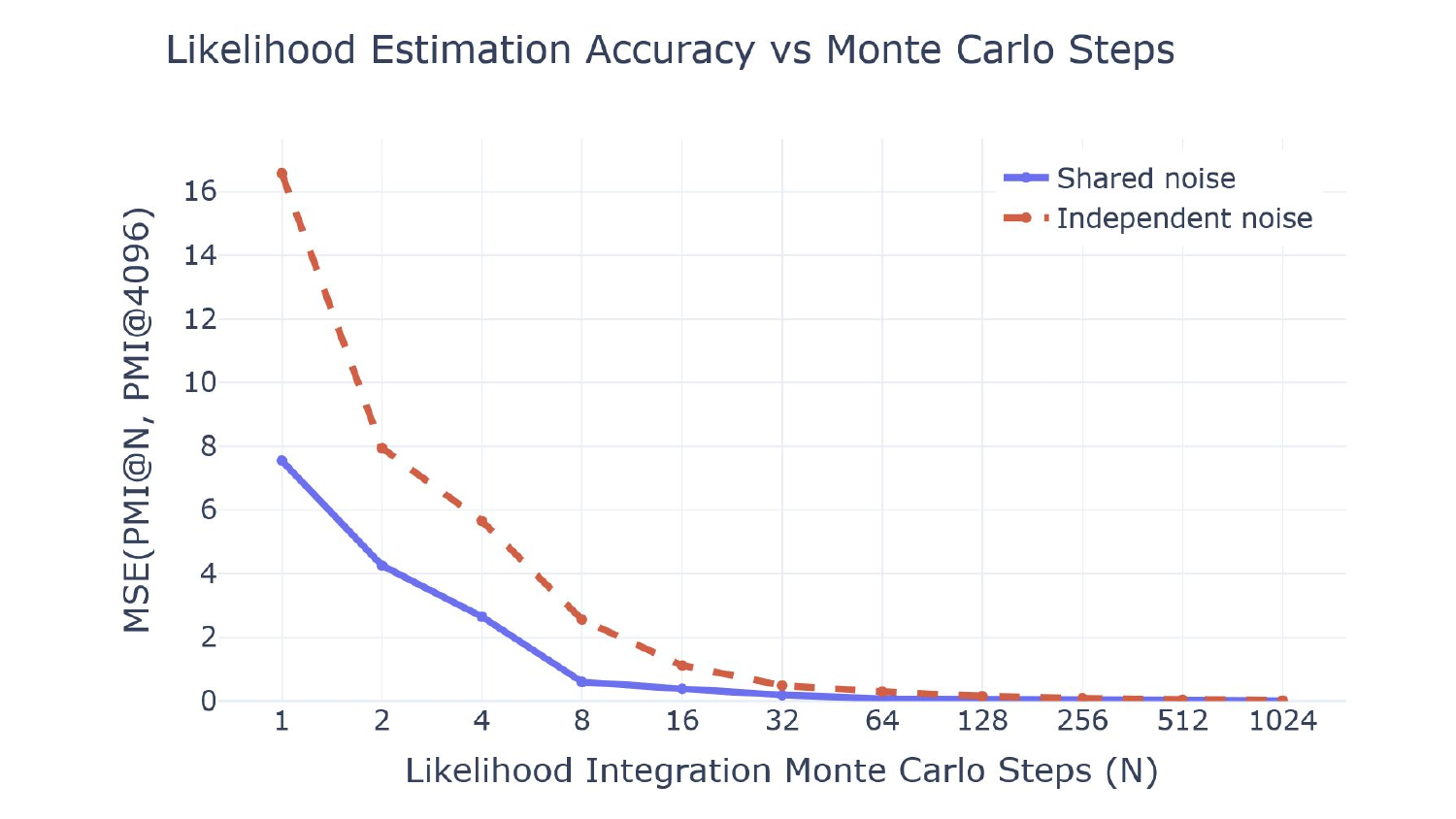}
    \caption{Monte Carlo PMI convergence versus number of Monte Carlo Integration Steps, measured as the MSE between the PMI when estimated with $N$ steps and the PMI when estimated with 4096 steps. Pixel space prediction, default configuration RePR-173M. Furthermore, sharing the noise among different likelihood calculations strongly reduces variance. We empirically choose $N=32$ steps at which diffusion integration uses $1.3\%$ of total flops during training.}
    \label{fig:likelihoodmontecarlo}
\end{figure}

\begin{table}[h]
\centering
\footnotesize
\setlength{\tabcolsep}{3pt}
\renewcommand{\arraystretch}{1}
\captionsetup{font=footnotesize,position=top,skip=4pt}
\begin{minipage}{0.8\textwidth}
\caption{Reconstruction performance for RePR-B Iter 3, 600ep, using the decoder configuration from RAE \citep{rae}.}
\label{tab:reconstruction}
\centering
\begin{tabular}{@{}lrr@{}}
\toprule
\textbf{Model} & \textbf{rFID} & \textbf{Reconstruction PSNR} \\
\midrule
DINOv2-B \citep{oquab2024dinov2} & 0.49 & 18.09 dB \\
RePR-B Iter. 3 600ep (Ours) & \textbf{0.47} & \textbf{21.72 dB} \\
\bottomrule
\end{tabular}
\end{minipage}
\end{table}


\newpage
\subsection{Implementation Details}
\label{app:implementation_details}
We use a network depth of $24$ layers, split as $18$ Encoder $E_\theta$ layers, and $6$ layers each for the Decoder $D_\theta$ and Representation output branch $R_\theta$. We split $R_\theta$ into three self-attention layers (identical to the encoder $E_\theta$) and three cross attention layers. With these design decisions, our architecture matches a standard ViT-L at inference at a width of $1024$, encoding $18$ $E_\theta$ and $6$ $R_\theta$ layers. 
 
To implement $E_\theta$ without providing absolute position information, we use RoPE position embeddings. For encoder register tokens, we assign RoPE position to the center of the unmasked tokens. When using center-relative offset queries (Sec.~\ref{sec:patch_features}), $R_\theta$ can output representations for unobserved patches, including patches outside of the image. Like convolutional networks, which are also spatially-equivariant, we find it empirically useful to pad one row and column of representation queries outside of the image. We find that $R_\theta$ emergently expresses low, but nonzero magnitude on these padding tokens. We explore adding a separate representation loss to the CLS tokens of $R_\theta$, which we find have better probing performance than pooled spatial features, similar to \citep{Caron_2021_ICCV}. We adapt all transformer architecture decisions from JiT \cite{li_jit}.

Because $\text{PMI}^*$ is asymmetric, and moreover, because PMI is not positive semi-definite, we indicate to $R_\theta$ which inputs correspond to incremental information $\rvz$ (Eq.~\ref{eq:repr_loss}) distinctly from $\rvx$ and $\rvy$. This added information asymmetrizes the dot product, allowing for perfect SVD reconstruction of the PMI at optimality. Asymmetrizing contrastive output heads has also been used to improve performance in prior work \citep{NEURIPS2022_494f876f}. To indicate the position of patch $\rvz$, we add a learned embedding to the patch features before encoding.

For all experiments, we report epochs as the total number of image views processed divided by the number of data samples in ImageNet. This keeps epochs meaningful with respect to FLOPs, invariant to input view repeating $A_\text{Repeat}$. To cheaply get learning rate decayed checkpoints for our models at different epoch counts, we train a single model with a stable learning rate that saves model checkpoints every 100 epochs, and we later linearly decay the learning rate for an additional 10 epochs from these checkpoints. For Figure~\ref{fig:compute_vs_jit}, we use a second iteration model trained for 420 total epochs, 210 for the first iteration and 210 for the second iteration, corresponding to $20\%$ and $35\%$ of total training FLOPs versus JiT-L at 200 epochs using RePR-44M and RePR-78M respectively. For Table~\ref{tab:jitbaseline}, we compare using a RePR-78M third iteration model trained for 830 epochs total, 210 each for the first and second iteration and 410 for the third iteration, corresponding to $25\%$ of total training FLOPs vs JiT-L at 600ep. For RAE comparisons (Table~\ref{tab:rae}), we use RePR-B third iteration trained for 1030 total epochs (210 Iter 1, 210 Iter 2, 610 Iter 3), using a total FLOPs of $2.19\times10^{20}$, fewer than the baseline I-JEPA-H with $2.45\times 10^{20}$ FLOPs. All FLOP results are reported using $6ND$, without self-attention.

Each Latent Forcing Evaluation takes roughly 31 total GPU hours on ImageNet using JiT-B, and 56 total GPU hours on GPIC using MiniT2I. We estimate that the results in Fig.~\ref{fig:scaling_laws} could be replicated on a single 8 GPU B200 node in 250 hours.

\subsection{Properties of PMI}
\label{app:pmi_properties}
\textbf{Reconstruction from PMI: }
With weak assumptions, PMI factorization can reconstruct the input. Let $\rvx$ be an uncorrupted view of an image and $\rvy$ be a pixel we wish to reconstruct. We assume nonzero support on all possible pixel values $\rvy$, which can be achieved by adding noise $\bepsilon \sim \mathcal{N}(0,\delta I)$ during training to pixels $\rvy$ where $\delta > 0$ may be arbitrarily small. Then, it is possible to reconstruct the data from the factorization of the PMI by iterating over all possible pixels $z$ and taking the maximum PMI with respect to $\rvx$: $\rvy = \arg\max_{\rvz \sim \text{All Pixels}} \langle R(\rvx),R(\rvz)\rangle$. Although the PMI factorization may reconstruct the input with arbitrary quality in theory, in practice reconstruction quality will be affected by the masking functions sampled during training and the convergence of the network. Empirically, RePR-B Iter 3. obtains a PSNR of 21.72 (Table~\ref{tab:reconstruction}), 3.63 dB higher than DINOv2 when using the RAE ViT-XL decoder \citep{rae}.

\begin{table}[h]
\small
\centering
\caption{Iteration 1 (Pixel) RePR configurations on ImageNet 256$\times$256.
Joint MAR and RePR steps (Alg.~\ref{alg:repr}) and MAR-only steps (Eq \ref{eq:mar_loss}) are interleaved with a 1:3 ratio, respectively.}
\label{tab:repr-imagenet-configs}
\begin{adjustbox}{max width=\linewidth}
\begin{tabular}{@{}lccccc@{}}
\toprule
Configuration & XT & T & S & B & L \\
\midrule
\multicolumn{6}{l}{\textbf{Shared settings}} \\
hidden dim & 256 & 384 & 512 & 768 & 1024 \\
inference parameters & 20M & 44M & 78M & 173M & 306M \\
heads & 4 & 6 & 8 & 12 & 16 \\
Encoder $E_\theta$ depth & \multicolumn{5}{c}{18} \\
patch size & \multicolumn{5}{c}{$16\times16$} \\
in-context tokens & \multicolumn{5}{c}{16} \\
batch size & \multicolumn{5}{c}{2048} \\
optimizer & \multicolumn{5}{c}{AdamW} \\
Adam $\beta_1$ & \multicolumn{5}{c}{0.9} \\
Adam $\beta_2$ & \multicolumn{5}{c}{0.95} \\
lr & \multicolumn{5}{c}{$2.4\times10^{-4}$} \\
$\mu$P weight lr multiplier & 3 & 2 & 1.5 & 1 & 0.75 \\
lr schedule & \multicolumn{5}{c}{warmup-stable-decay} \\
warmup epochs & \multicolumn{5}{c}{20} \\
decay epochs & \multicolumn{5}{c}{10} \\
weight decay & \multicolumn{5}{c}{0.0} \\
dropout & \multicolumn{5}{c}{0.0} \\
random resized crop scale & \multicolumn{5}{c}{$[0.8,1.0]$} \\
horizontal flip & \multicolumn{5}{c}{enabled} \\
mask strategy & \multicolumn{5}{c}{block} \\
minimum mask ratio $M_A$ & \multicolumn{5}{c}{75\% (64 Visible Patches)} \\
minimum mask ratio $M_B$ & \multicolumn{5}{c}{93.75\% (16 Visible Patches)} \\
\midrule
\multicolumn{6}{l}{\textbf{MAR hyperparameters}} \\
Decoder $D_\theta$ depth & \multicolumn{5}{c}{6} \\
Denoising MLP $M_\theta$ depth & \multicolumn{5}{c}{3} \\
mask repeats & \multicolumn{5}{c}{1} \\
noise repeats & \multicolumn{5}{c}{4} \\
logit-normal $\mu$ & \multicolumn{5}{c}{$1.2$} \\
logit-normal $\sigma$ & \multicolumn{5}{c}{1.2} \\
uniform mixture weight & \multicolumn{5}{c}{10\%} \\
\midrule
\multicolumn{6}{l}{\textbf{RePR hyperparameters}} \\
$R_\theta$ encoder depth (First 18 layers are $E_\theta)$ & \multicolumn{5}{c}{21} \\
$R_\theta$ decoder depth & \multicolumn{5}{c}{3} \\
Output dimension & \multicolumn{5}{c}{1024} \\
$p_\text{Repeat}$ & \multicolumn{5}{c}{16} \\
$A_\text{Repeat}$ & \multicolumn{5}{c}{8} \\
Monte Carlo steps $N$ & \multicolumn{5}{c}{32} \\
$f(t)$ logit-normal $\mu$ (pixel prediction) & \multicolumn{5}{c}{1.2} \\
$f(t)$ logit-normal $\mu$ (feature prediction) & \multicolumn{5}{c}{1.6} \\
$f(t)$ logit-normal $\sigma$ & \multicolumn{5}{c}{1.2} \\
$g(t)$  & \multicolumn{5}{c}{$(1-t)/\max(0.33,t)^2$} \\
RePR loss weight & \multicolumn{5}{c}{0.005} \\
RePR grad clip & \multicolumn{5}{c}{0.2} \\
Padding dropout & \multicolumn{5}{c}{0.01} \\
latent normalization (feature prediction) & \multicolumn{5}{c}{$\mu=0, \sigma^2=0.5^2$} \\
\bottomrule
\end{tabular}
\end{adjustbox}
\par\smallskip
\end{table}

\begin{table}[h!]
\small
\centering
\caption{Latent Forcing \citep{baade2026latent} decoder training at 256$\times$256 resolution with
16$\times$16 image patches. JiT-B used for scaling laws, JiT-L used for baseline comparisons.}
\label{tab:decoder-training}
\begin{tabular}{@{}lccc@{}}
\toprule
config & ImageNet & ImageNet & GPIC \\
\midrule
decoder & JiT-B \citep{li_jit} & JiT-L \citep{li_jit}  & MiniT2I \citep{minit2i2026} \\
hidden dim & 768 & 1024 & 768 \\
backbone depth & 12 & 20 & 17 \\
output experts \cite{baade2026latent} & 2 $\times$ 1024 dim & 2 $\times$ 1024 dim & - \\
text encoder & - & - & FLAN-T5-large \\
max text tokens & - & - & 128 \\
\midrule
epochs & \multicolumn{3}{c}{80} \\
global batch size & \multicolumn{3}{c}{1024} \\
optimizer & \multicolumn{3}{c}{AdamW} \\
peak lr & \multicolumn{3}{c}{$2\times10^{-4}$} \\
lr schedule & \multicolumn{3}{c}{constant after warmup} \\
warmup epochs & \multicolumn{3}{c}{5} \\
weight decay & \multicolumn{3}{c}{0.0} \\
\midrule
pixel time $\mu$ & \multicolumn{3}{c}{$0.4$} \\
pixel time $\sigma$ & \multicolumn{3}{c}{0.8} \\
latent time $\mu$ & \multicolumn{3}{c}{$1.6$} \\
latent time $\sigma$ & \multicolumn{3}{c}{1.0} \\
latent selection probability & \multicolumn{3}{c}{0.5} \\
latent loss weight & \multicolumn{3}{c}{1.0} \\
$t_{\min}$ \citep{li_jit} & \multicolumn{3}{c}{0.05} \\
latent normalization & \multicolumn{3}{c}{$\mu=0, \sigma^2=0.5^2$} \\
\bottomrule
\end{tabular}
\end{table}

\clearpage
\subsection{Qualitative Results}
We present \emph{uncurated} results of our approach on ImageNet. We follow the exact CFG-Interval+AutoGuidance Setup from Latent Forcing \citep{baade2026latent}, applying CFG $3.0$ to latent diffusion on $[0.06,1.0]$ and AutoGuidance with $w=1.5$. FLOP calculations include AutoGuidance, which uses $1.6\%$ of total flops during Training

\begin{figure}[h]
    \centering
    \includegraphics[width=0.9\textwidth, trim={0.0cm 0cm 0.0cm 0.0cm}, clip]{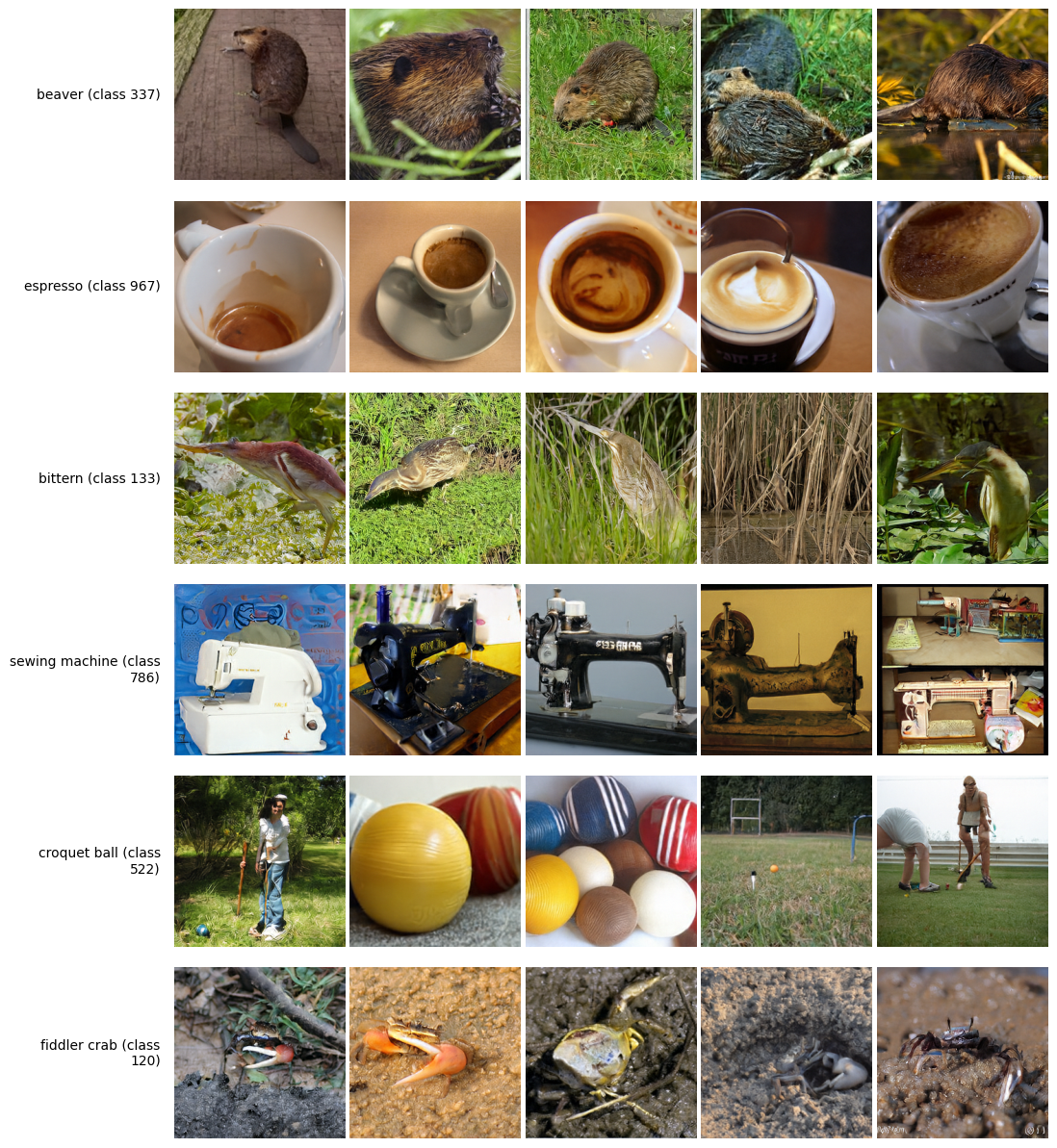}
    \caption{\emph{Uncurated} class-conditional Imagenet Generations, from our compute optimal JiT-L.}
    \label{fig:uncurated_imagenet}
\end{figure}

\begin{figure}[h]
    \centering
    \includegraphics[width=0.5\textwidth, trim={0.0cm 0cm 0.0cm 0.0cm}, clip]{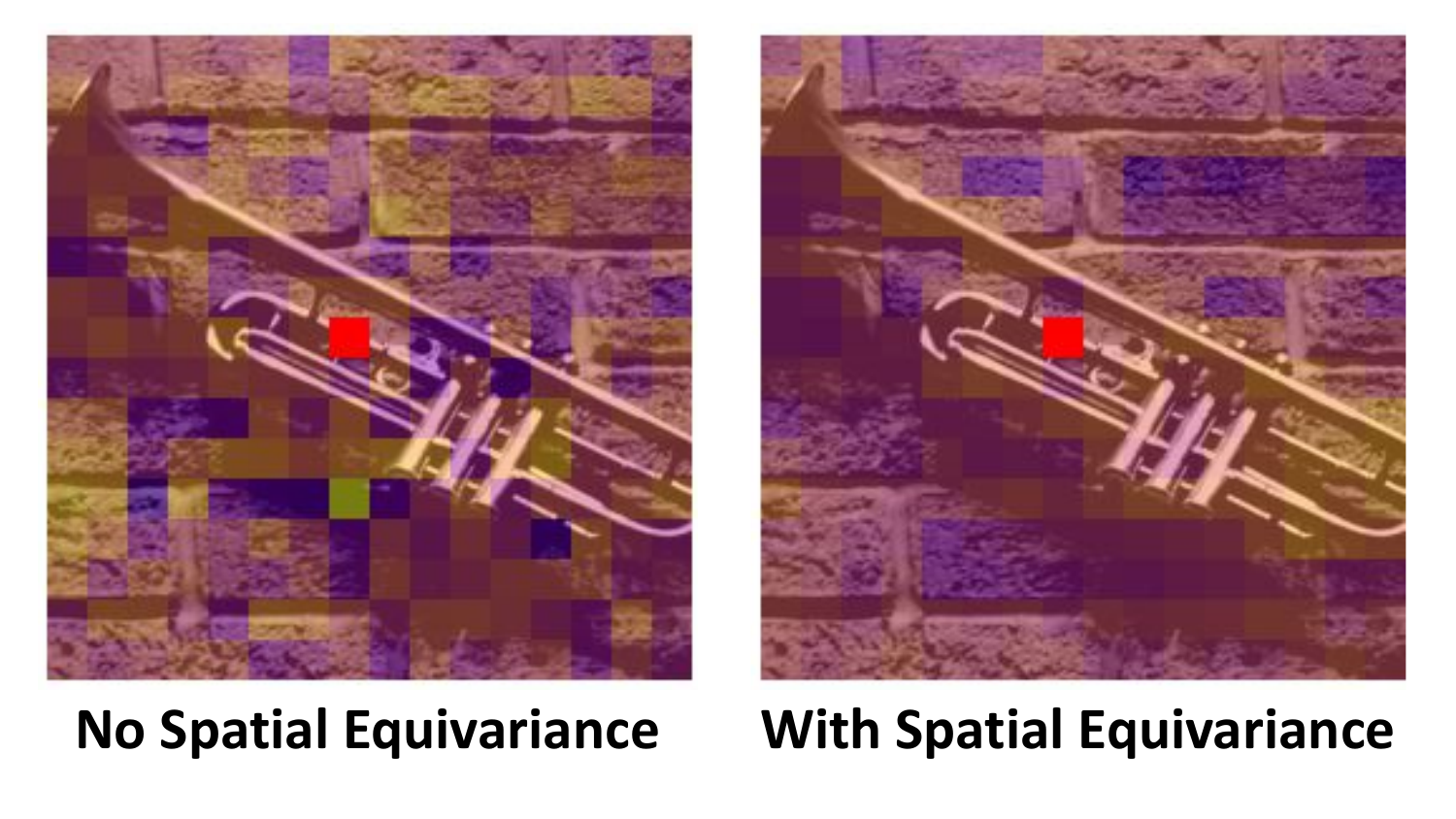}
    \caption{Feature cosine similarity between patch features and a center patch feature shown in red, with and without spatial equivariance (Sec~\ref{sec:patch_features}).}
    \label{fig:uncurated_imagenet}
\end{figure}

\clearpage
\subsection{Watermarking}

\begin{figure}[h]
    \centering
    \includegraphics[width=0.9\textwidth, trim={0.0cm 1.5cm 0.0cm 1.5cm}, clip]{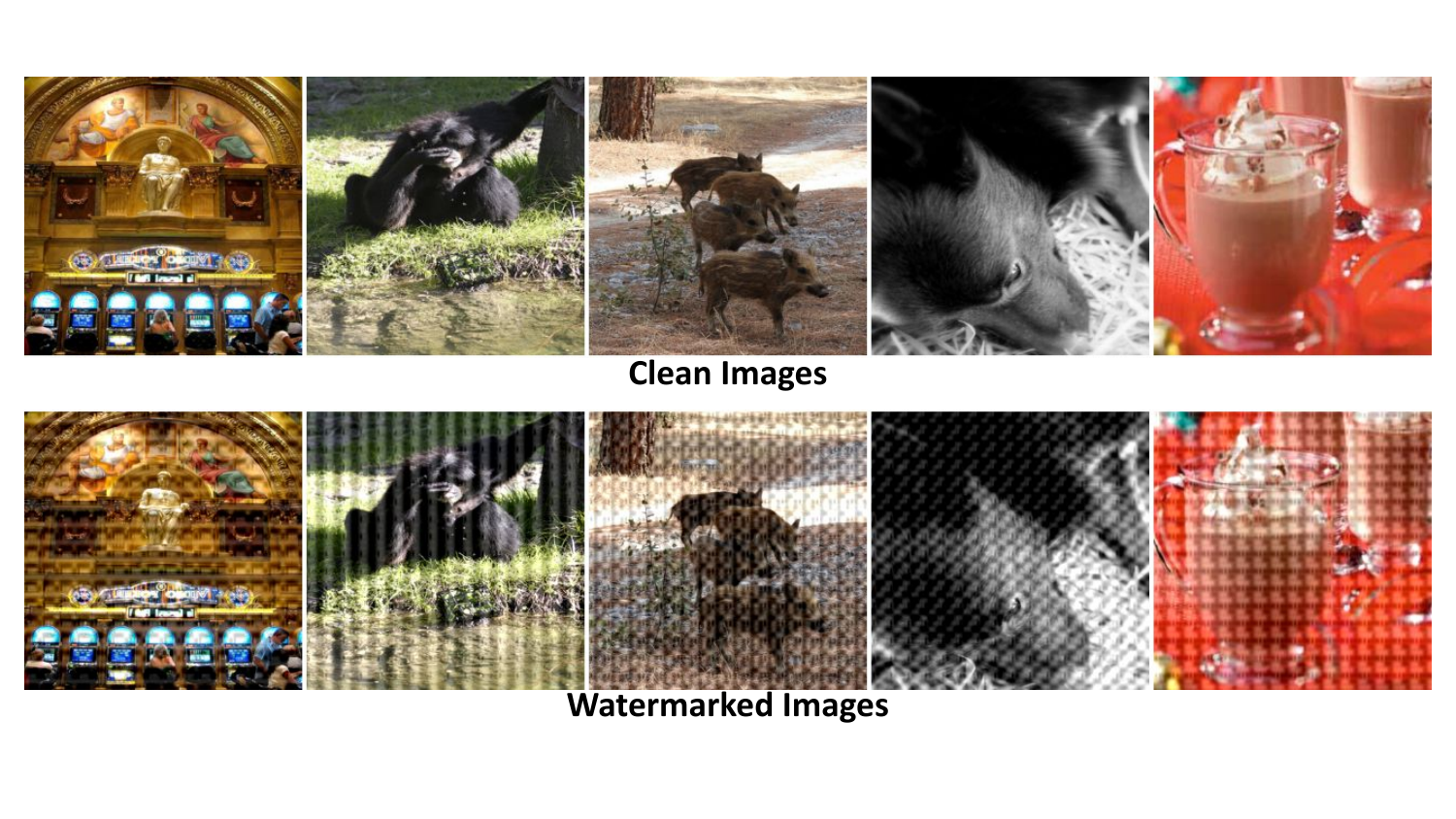}
    \caption{Example Clean and Watermarked images from Imagenet, used in Sec.~\ref{sec:feature_suppression}. We follow the watermarking strategy from \cite{enigma2026obsessedencoder} exactly, except for that we shrink watermarks from $32\times 32$ to $16\times16$ pixels, so that every individual patch provides sufficient signal for suppression.}
    \label{fig:uncurated_imagenet}
\end{figure}

\clearpage
\subsection{Raw Numbers for Scaling Plots}
\begin{table}[h!]
\small
\centering
\caption{ImageNet scaling result tables, evaluated using FID-10K after 80 epochs of generative model training. Iteration 2 epoch counts include 210 epochs of Iteration 1 training (200 epochs + 10 epochs of learning rate decay). The DINOv2 baseline obtains 17.95 FID at $1.1685129\times10^{22}$ encoder-training FLOPs.}
\label{tab:imagenet-scaling-raw}
\begin{tabular}{@{}lrrrr@{}}
\toprule
& \multicolumn{2}{c}{Iteration 1: Pixel} & \multicolumn{2}{c}{Iteration 2: Latent} \\
\cmidrule(lr){2-3}\cmidrule(l){4-5}
Encoder (width) & Epochs & FID & Epochs & FID \\
\midrule
20M (256) & 110 & 41.61 & $210+110$ & 36.24 \\
          & 210 & 40.19 & $210+210$ & 34.89 \\
          & 410 & 37.01 & -- & -- \\
          & 610 & 37.21 & -- & -- \\
\midrule
44M (384) & 110 & 40.93 & $210+110$ & 31.64 \\
          & 210 & 38.20 & $210+210$ & 29.48 \\
          & 410 & 36.07 & $210+410$ & 29.22 \\
          & 610 & 34.82 & -- & -- \\
\midrule
78M (512) & 110 & 38.26 & $210+110$ & 29.81 \\
          & 210 & 35.72 & $210+210$ & 27.80 \\
          & 410 & 34.31 & $210+410$ & 27.36 \\
          & 610 & 34.00 & -- & -- \\
\midrule
173M (768) & 110 & 37.22 & $210+210$ & 25.82 \\
           & 210 & 33.73 & $210+410$ & 24.73 \\
           & 410 & 31.27 & -- & -- \\
           & 610 & 30.68 & -- & -- \\
\midrule
306M (1024) & 110 & 36.14 & -- & -- \\
            & 211 & 31.23 & -- & -- \\
\bottomrule
\end{tabular}
\end{table}

\begin{table}[h!]
\small
\centering
\caption{GPIC scaling result tables, evaluated using FD-DINOv2-10K after 1 GPIC epoch of generative model training. For consistency between approaches, epochs are measured as effective epochs on ImageNet. For reference, 80 ImageNet epochs is roughly equal to 1 GPIC epoch. The WebSSL-DINO-1B reference obtains an FD-10K of 277.1 at $9.78\times10^{21}$ encoder-training FLOPs. MiniT2I obtains an FD-10K of 546.3.}
\label{tab:gpic-scaling-raw}
\begin{tabular}{@{}lrrrr@{}}
\toprule
& \multicolumn{2}{c}{Iteration 1: Pixel} & \multicolumn{2}{c}{Iteration 2: Latent} \\
\cmidrule(lr){2-3}\cmidrule(l){4-5}
Encoder (width) & ImageNet Epochs-Worth & FD & ImageNet Epochs-Worth & FD \\
\midrule
44M (384) & 110 & 514 & $210+110$ & 416 \\
          & 210 & 479 & $210+210$ & 393 \\
          & 410 & 456 & -- & -- \\
\midrule
78M (512) & 110 & 498 & $210+110$ & 384 \\
          & 210 & 463 & $210+210$ & 365 \\
          & 410 & 443 & -- & -- \\
\midrule
173M (768) & 110 & 450 & $210+110$ & 359 \\
           & 210 & 423 & $210+210$ & 354 \\
           & 410 & 409 & -- & -- \\
\midrule
306M (1024) & 110 & 451 & $210+110$ & 347 \\
            & 210 & 404 & $210+210$ & 325 \\
            & 410 & 381 & -- & -- \\
\bottomrule
\end{tabular}
\end{table}

Power-law fits predict DINOv2-level ImageNet FID at
$1.504\times10^{21}$ encoder-training FLOPs, $1/7.77$ of the compute, and WebSSL-level GPIC FD at $1.229\times10^{21}$ FLOPs, $1/7.96$ of the compute.

\end{document}